\documentclass{article}
\usepackage[english]{babel}
\usepackage[utf8]{inputenc}
\usepackage{bm}
\usepackage{amsmath,amsthm,amsfonts}
\usepackage{graphicx}
\usepackage{color}
\usepackage{multirow}
\usepackage{subcaption}
\usepackage{fancyhdr}  % 引入 fancyhdr 宏包
\usepackage{geometry}  % 用于设置页面几何

\graphicspath{ {./figs/} }

\date{}

\newcommand{\mytitle}{A Hybrid Framework for Spatial Interpolation: \\ Merging Data-driven with Domain Knowledge}

\title{\mytitle}

\author{
Cong Zhang \textsuperscript{1}
\and
Shuyi Du \textsuperscript{1,2}
\and
Hongqing Song \textsuperscript{1,2,}\thanks{Correspondence: {\tt songhongqing@ustb.edu.cn}}
\and
Yuhe Wang \textsuperscript{1,3,}\thanks{Correspondence: {\tt yuhe.wang@tamu.edu}}
}

\date{\normalsize \today}

\begin{document}

\maketitle

\footnotetext[1]{National \& Local Joint Engineering Lab for Big Data Analysis and Computing Technology, Beijing 100190, China.}
\footnotetext[2]{School of Civil and Resource Engineering, University of Science and Technology Beijing, Beijing 100083, China.}
\footnotetext[3]{Institute for Scientific Computation, Texas A\&M University, College Station, Texas 77843, USA.}

\thispagestyle{fancy}

\begin{abstract}
Estimating spatially distributed information through the interpolation of scattered observation datasets often overlooks the critical role of domain knowledge in understanding spatial dependencies. Additionally, the features of these datasets are typically limited to the spatial coordinates of the scattered observation locations. In this paper, we propose a hybrid framework that integrates data-driven spatial dependency feature extraction with rule-assisted spatial dependency function mapping to augment domain knowledge. We demonstrate the superior performance of our framework in two comparative application scenarios, highlighting its ability to capture more localized spatial features in the reconstructed distribution fields. Furthermore, we underscore its potential to enhance nonlinear estimation capabilities through the application of transformed fuzzy rules and to quantify the inherent uncertainties associated with the observation datasets. Our framework introduces an innovative approach to spatial information estimation by synergistically combining observational data with rule-assisted domain knowledge. 

\end{abstract}

\section{Introduction}

Spatially dependent properties are widespread across various fields including subsurface resource exploitation \cite{yarus2006practical,mohammadpour2024machine,wang2020generalized,yan2018enhanced,mi2017enhanced}, water resources management \cite{li2008review,yavuz2012spatial}, traffic engineering \cite{shamo2015linear,lowry2014spatial}, and environmental studies \cite{hong2005spatial,li2022machine}. These spatially varying attributes are typically recorded at a limited number of observed points or monitoring stations, often insufficient to represent the entire and multiscale spatial distribution accurately \cite{zhang2019novel,lengyel2023modelling,mi2017enhanced}. Due to the impracticality of monitoring every part of a domain to fully capture its attribute field, spatial interpolation is commonly employed to estimate values at unobserved locations, utilizing observed data and underlying spatial dependency correlations \cite{jensen2000statistics,masoudi2024analysing}.

Numerous spatial interpolation methods exist, as reviewed by Li and Heap \cite{li2014spatial}, who suggest that these methods often rely on similar principles, such as inverse distance weighting \cite{shepard1968two} and Kriging \cite{matheron1963principles}. The inverse distance weighting approach, which is built on a straightforward distance-based method, is generally unsuitable for spatially aggregated observations. Kriging, on the other hand, functions as an unbiased linear estimator under the assumption of a stationary normal distribution. Since it utilizes the conditional distribution of a Gaussian process, Kriging is essentially a specific case of Gaussian process regression \cite{chen2020deepkriging,wang2019nearest}. However, all these methods are founded on simplified assumptions, including stationarity, linearity, and normality, which limit their applicability in real-world scenarios \cite{wang2019nearest,shi2021non,zhu2020spatial}. In particular, their reliance on linear interpolation makes them inadequate for handling complex nonlinear spatial dependency relationships, often resulting in suboptimal estimations for continuous attribute fields. Therefore, there is a clear need for alternative approaches that can construct non-stationary, nonlinear, and non-Gaussian spatial dependency functions for spatial interpolation.

Given their versatile nonlinear approximation capabilities, the emerging data-driven approaches are promising to provide an alternative framework for spatial information modeling \cite{li2022machine,kadow2020artificial,du2020connectivity,du2023novel,zhang2019potential}. Various machine learning-based methods, such as random forests \cite{sekulic2021high,han2024spatial,geerts2024georf}, artificial neural networks \cite{korjani2016reservoir,vsapina2016comparison,sunayana2020use}, radial basis function networks \cite{shi2021non,liang2024study}, long short-term memory networks \cite{otake2020deep,yu2024spatial}, convolutional neural networks \cite{hashimoto2020sicnn,suto2021image}, conditional generative adversarial networks \cite{zhu2020spatial,yan2024conditional,rakotonirina2024spatial}, and ensemble learning \cite{egana2021ensemble,papacharalampous2024uncertainty}, have been successfully adapted and applied to spatial interpolation. However, without significant reengineering, these original machine learning methods are not inherently designed for spatial interpolation and often fail to account for the available spatial configuration information. In other words, directly applying these machine learning models may pose challenges and may not be suitable for most spatial interpolation scenarios \cite{sekulic2020random,nwaila2024spatial}.

While methods involving convolutional neural networks and generative adversarial networks are well-suited for imagery or regularly gridded datasets, such as terrain elevation images, observational data are often irregularly sparse and scattered \cite{stoica2010new,beiser2024comparison}. Traditional deep learning methods may not be ideal for extracting spatial dependency features from such sparse observations, as they typically only rely on the spatial coordinates of the observation locations as input features \cite{tsagkatakis2019survey}. To address this limitation, modified approaches have been developed. For example, Wu et al. proposed an inductive graph neural network Kriging model that better leverages distance information \cite{wu2021inductive}, while Ma et al. introduced a geo-layer into long short-term memory networks to integrate spatial correlations from monitoring stations \cite{ma2019temporal}. In essence, effectively extracting spatial dependency features from irregularly sparse observed data necessitates considering both individual observations and their neighboring spatial configurations.

In addition to the data-driven approaches, the family of rule-based fuzzy inference systems can effectively model nonlinear functions by leveraging inherent reasoning paradigms \cite{jang1993anfis, mamdani1974application, mamdani1975experiment,mao2022decision,guerra2024survey}. Unlike data-driven methods, fuzzy inference systems can directly and broadly incorporate expert knowledge, making them valuable for enhancing spatial interpolation \cite{mao2022decision}. For example, the first law of geography, which states that “near things are more related than distant things” \cite{tobler1970computer}, can be articulated as a fuzzy IF-THEN rule—IF two spatial observations are closer, THEN their spatial dependency correlations are stronger. This rule underpins many existing interpolation methods that rely on concepts of distance and neighborhood. Similarly, there is a wealth of domain-specific knowledge related to both general spatial dependencies and particular interpolation scenarios. For instance, Yesilkanat et al. \cite{yecsilkanat2017spatial} demonstrated the successful application of domain expertise for spatial interpolation of environmental radioactivity using fuzzy IF-THEN rules, either collected or self-defined. However, their fuzzy rulesets were tailored to specific domains rather than forming a general rule base for spatial dependency extraction. Therefore, there is a need to develop a more general rule-based spatial interpolation framework that can effectively exploit and utilize latent domain knowledge. It is important to note that domain knowledge related to spatial dependency is often difficult to collect and represent in the form of fuzzy IF-THEN rules. Moreover, the lack of standardized approaches to transforming domain knowledge into rule bases limits the effective use of such knowledge. Consequently, it is both necessary and advantageous to systematically and automatically generate fuzzy rulesets to incorporate domain knowledge. Another advantage of rule-based fuzzy inference systems is their ability to tolerate inaccurate information. Observation data and monitoring records are often subject to errors from various sources \cite{bayat2021uncertainty}, which can negatively impact interpolation accuracy. A robust spatial interpolation approach should account for these uncertainties associated with the observed data. Notably, fuzzy logic is well-suited for handling such inherent uncertainties, thereby enhancing the performance of spatial interpolation \cite{tapoglou2014spatio,zhao2020reservoir}.

In this paper, we present a hybrid framework merging data-driven and domain knowledge for interpolating spatially dependent properties. The framework extracts latent spatial dependency basis by leveraging observation data and neighboring information. By embedding a fuzzy inference system within our adaptive network, we can automatically transform domain knowledge into fuzzy rulesets. This framework capitalizes on the advantages of fuzzy reasoning, particularly its ability to tolerate inaccurate information and manage nonlinear spatially dependent functions. We validate our framework in two scenarios: subsurface formation parameter estimation and air quality mapping. Additionally, we conduct comparative studies to assess the estimation performance quantitatively and qualitatively against conventional interpolation techniques, including ordinary Kriging, inverse distance weighting, and Gaussian process regression.

\section{Hybrid data-driven and rule-assisted learning framework}

\begin{figure}[h!]
\centering
\includegraphics[width=0.8 \textwidth]{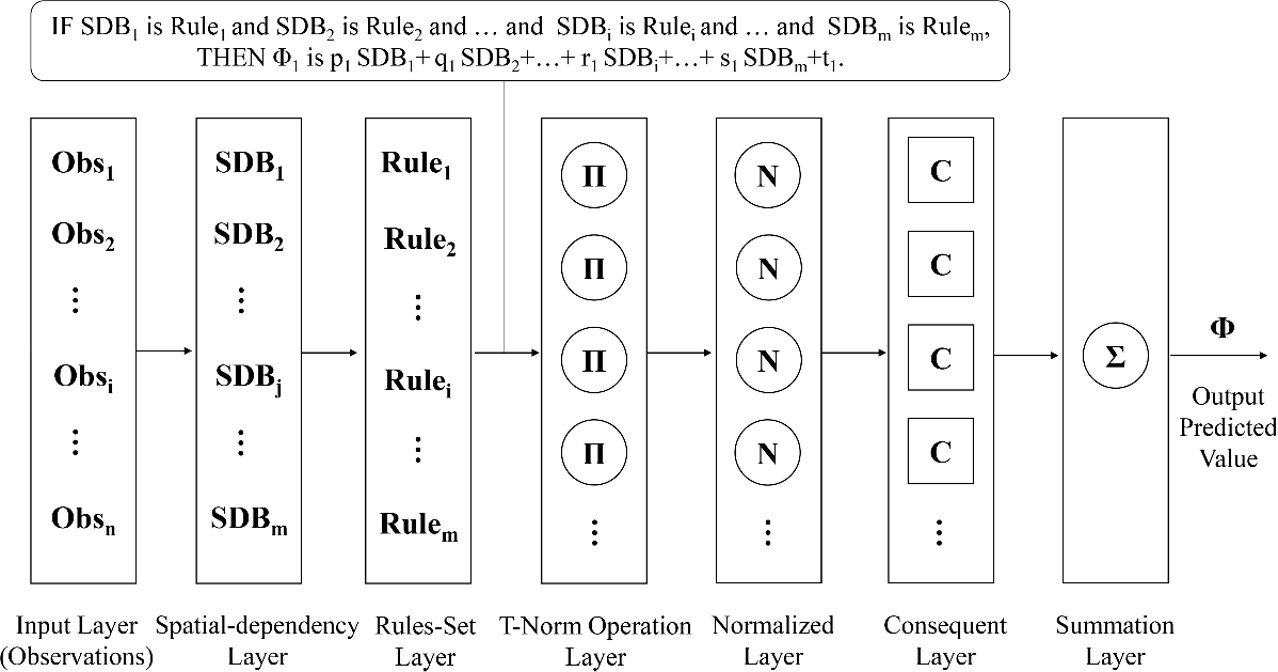}
\caption{\textbf{The network architecture of the hybrid framework based on ANFIS}}
\label{fig:network-1}
\end{figure}

To address the limitations mentioned above, our framework combines the extraction of spatial dependency features from observation data with the transformation of latent domain knowledge into fuzzy rulesets. For constructing a fuzzy inference system, it is ideal to automatically convert domain knowledge into fuzzy IF-THEN rules. The Adaptive-Network-based Fuzzy Inference System (ANFIS) developed by J.S. Jang \cite{jang1993anfis} provides a solid foundation for this task. However, the number of rules generated by ANFIS can become intractable as the number of input features increases. Specifically, if the number of membership functions assigned to each input dimension is fixed and the dimensionality increases linearly, the number of generated rules grows exponentially. This exponential growth can significantly limit the applicability of ANFIS when incorporating both spatial coordinates and relevant neighboring information as input features. Inspired by ANFIS, we propose an enhanced architecture to overcome this issue by input feature decomposition. As shown in \textbf{Figure \ref{fig:network-1}}, our architecture integrates a data-driven approach with rule-based assistance. The network consists of an intrinsic input layer, a ruleset layer, a T-Norm operation layer, a normalization layer, a consequent layer, and a summation layer. Additionally, it includes a newly designed spatial dependency layer to exploit sparse observation data and their neighboring information. Essentially, our architecture retains the ANFIS learning mechanism for constructing fuzzy IF-THEN rules and approximating the estimation function.

\begin{figure}[h!]
\centering
\includegraphics[width=0.8 \textwidth]{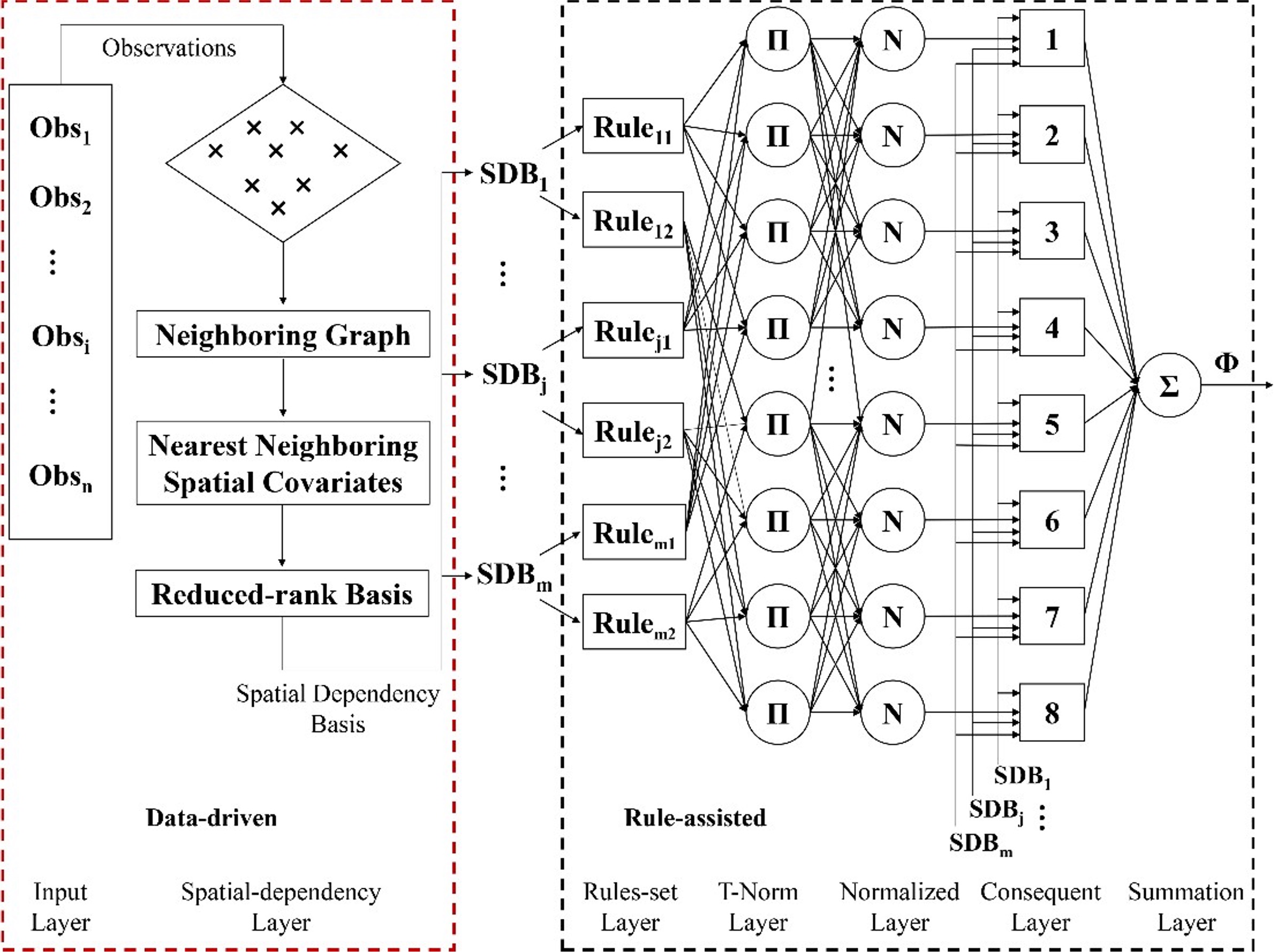}
\caption{\textbf{The detailed architecture of the hybrid framework}}
\label{fig:network-2}
\end{figure}

The detailed architecture of our hybrid framework is illustrated in \textbf{Figure \ref{fig:network-2}}. It has two main components: 1) data-driven Spatial Dependency Basis (SDB) extraction, and 2) rule-assisted spatial dependency function approximation. The SDB serves as a critical link between these two components, significantly reducing the number of generated rules. In the following sections, we outline the main steps and their corresponding implementation.

\subsection{Spatial Dependency Basis (SDB) extraction}

We apply the reduced-rank approach for obtaining SDB \cite{amato2020novel}. The SDB is extracted from spatial observations to represent fixed spatial features. Rather than using the direct input features from these spatial observations, we use SDB as the input for the rule-based ANFIS. This approach aims to minimize the number of automatically generated fuzzy IF-THEN rules while fully leveraging the available spatial observations. As illustrated in \textbf{Figure \ref{fig:network-2}}, the extraction of the SDB begins with constructing nearest neighboring spatial covariates.

\subsubsection{Nearest neighboring spatial covariates}

We use nearest neighboring spatial covariates \cite{sekulic2021high,sekulic2020random}, which is the combination of the environmental covariates with the spatial covariates, to comprehensively describe the spatial dependency relationship from spatial observations. Compared to approaches that consider only spatial coordinates as input features, these combined covariates better characterize spatial correlations by fully accounting for spatial coordinates, observation data, and neighboring configurations. The neighboring configurations are established by constructing a neighboring graph for each observation location. We use nearest neighboring algorithm to select the \textit{m} nearest neighbors for each observation location based on Euclidean distance, see Equation \eqref{eq1}. 

\begin{equation}
\label{eq1}
d_{ij} = \| \mathbf{Obs}_i - \mathbf{Obs}_j \|_2
\end{equation}

where $\mathbf{Obs}_i$ and $\mathbf{Obs}_j$ are the $i$-th and $j$-th observed points, respectively. In 2D, $\mathbf{Obs}_i = (x_i, y_i)$, and in 3D, if necessary, $\mathbf{Obs}_i = (x_i, y_i, z_i)$. The value $d_{ij}$ represents the Euclidean distance between $\mathbf{Obs}_i$ and $\mathbf{Obs}_j$.

\begin{figure}[h!]
\centering
\includegraphics[width=0.5 \textwidth]{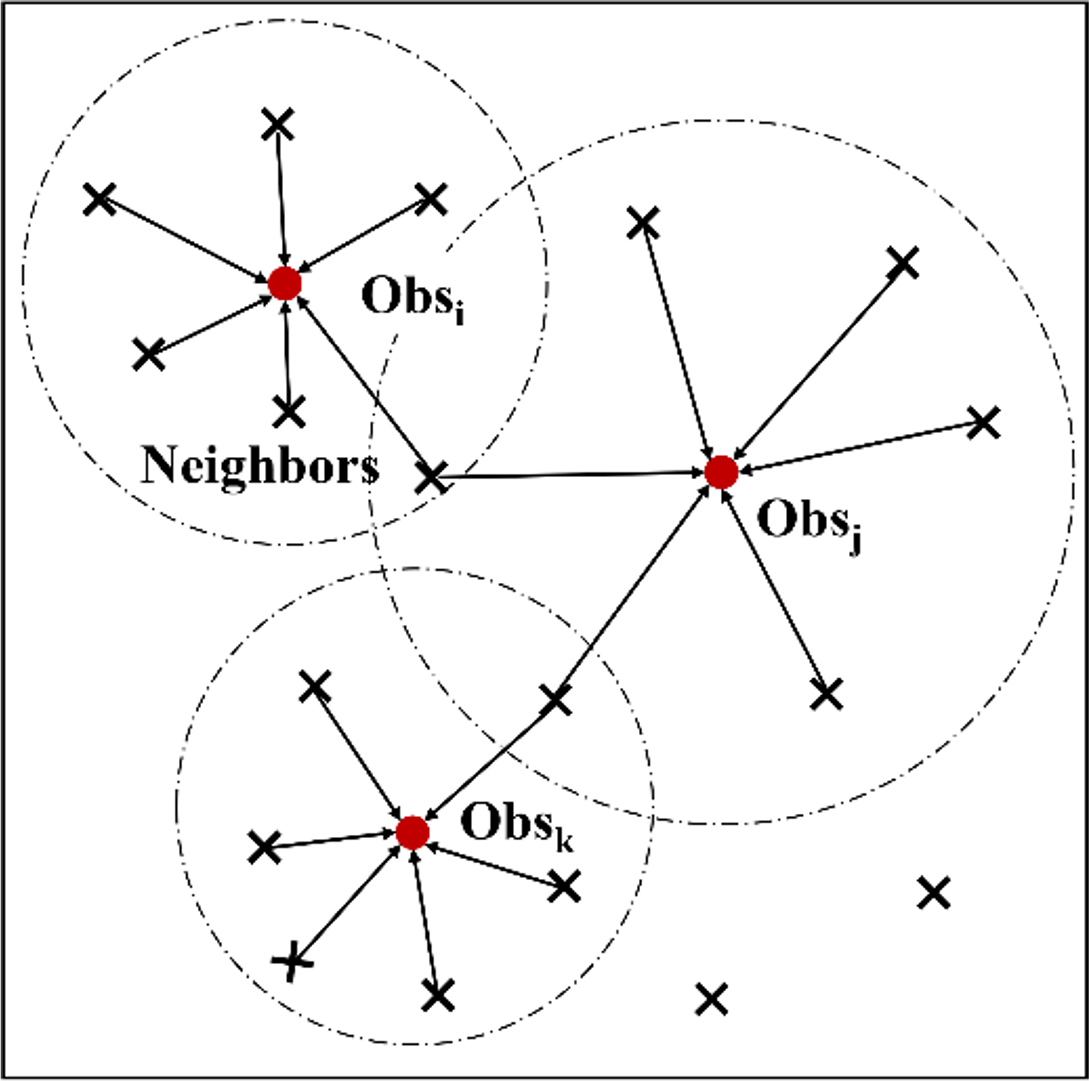}
\caption{\textbf{An illustration of Neighboring graph}}
\label{fig:neighbor}
\end{figure}

As illustrated in \textbf{Figure \ref{fig:neighbor}}, the observation locations and their corresponding neighbors form a neighboring graph. This graph is then used to construct the associated features, allowing the features at a specific observation location to encompass both its own data and the information from its neighbors.

We can express the neighboring graph using nearest neighboring spatial covariates to generate more valuable features for each observation. For a given observation $\text{Obs}_i = (x_i, y_i)$ with observation value of $\Phi_i$ and the $m$ nearest neighbors, the corresponding neighboring spatial covariates are written as:

\begin{equation}
\text[{Obs}_i] = [x_i, y_i, x_i^1, y_i^1, \Phi_i^1, x_i^2, y_i^2, \Phi_i^2, \dots, x_i^j, y_i^j, \Phi_i^j, \dots, x_i^m, y_i^m, \Phi_i^m]_{1 \times (3m+2)}
\end{equation}

where $x_i$ and $y_i$ are the spatial coordinates of the $i$-th observed point, and $i = 1, 2, \dots, N$ if there are $N$ observations in total. For the case with $N$ observed points, $\text{Obs}_1, \text{Obs}_2, \dots, \text{Obs}_i, \dots, \text{Obs}_N$, we can construct a matrix of nearest neighboring spatial covariates as shown in Equation \eqref{eq3}. The superscript $j$ represents the $j$-th nearest neighbor of the $i$-th observed point, and $j = 1, 2, \dots, m$.

\begin{equation}
\label{eq3}
\resizebox{0.93\textwidth}{!}{  % 自动缩放至页面宽度
$
\begin{bmatrix}
\mathit{Obs}_1 \\
\mathit{Obs}_2 \\
\vdots \\
\mathit{Obs}_i \\
\vdots \\
\mathit{Obs}_N \\
\end{bmatrix}
=
\left[
\begin{array}{cccccccccccccccc}
x_1 & y_1 & x_1^1 & y_1^1 & \Phi_1^1 & x_1^2 & y_1^2 & \Phi_1^2 & \dots & x_1^j & y_1^j & \Phi_1^j & \dots & x_1^m & y_1^m & \Phi_1^m \\
x_2 & y_2 & x_2^1 & y_2^1 & \Phi_2^1 & x_2^2 & y_2^2 & \Phi_2^2 & \dots & x_2^j & y_2^j & \Phi_2^j & \dots & x_2^m & y_2^m & \Phi_2^m \\
\vdots & \vdots & \vdots & \vdots & \vdots & \vdots & \vdots & \vdots & \ddots & \vdots & \vdots & \vdots & \ddots & \vdots & \vdots & \vdots \\
x_i & y_i & x_i^1 & y_i^1 & \Phi_i^1 & x_i^2 & y_i^2 & \Phi_i^2 & \dots & x_i^j & y_i^j & \Phi_i^j & \dots & x_i^m & y_i^m & \Phi_i^m \\
\vdots & \vdots & \vdots & \vdots & \vdots & \vdots & \vdots & \vdots & \ddots & \vdots & \vdots & \vdots & \ddots & \vdots & \vdots & \vdots \\
x_N & y_N & x_N^1 & y_N^1 & \Phi_N^1 & x_N^2 & y_N^2 & \Phi_N^2 & \dots & x_N^j & y_N^j & \Phi_N^j & \dots & x_N^m & y_N^m & \Phi_N^m
\end{array}
\right]
$
}
\end{equation}

For each observation location, the dimension of its nearest neighbor spatial covariates is $3m+2$. Assigning two membership functions to each dimension results in the adaptive neural networks automatically generating $2^{3m+2}$ fuzzy IF-THEN rules. This can lead to a significant number of generated rules, causing computational challenges in tuning and updating the hyperparameters during training to accurately map the nonlinear spatial dependency function. To address this, we further extract primary features that can characterize the spatial correlations using SDB.

\subsubsection{Reduced-rank basis extraction}

We use Uniform Manifold Approximation and Projection (UMAP) \cite{mcinnes2018umap} to decompose the nearest neighboring spatial covariates into several fixed reduced-rank bases specifically targeting spatial dependency basis. According to Equation \eqref{eq3}, the dimension of the nearest neighboring Spatial Covariates (SC) is $3m+2$. If we want to reduce the dimensionality to $d$, we can write such reduction as in Equations \eqref{eq4} and \eqref{eq5} to extract the Spatial Dependency Basis (SDB).  

\begin{equation}
\label{eq4}
\resizebox{0.93\textwidth}{!}{  % 自动缩放至页面宽度
$    
    SC_i = \mathit{Obs}_i = 
\left[
\begin{array}{cccccccccccc}
    x_i & y_i & x_i^1 & y_i^1 & \Phi_i^1 & x_i^2 & y_i^2 & \Phi_i^2 & \dots & x_i^m & y_i^m & \Phi_i^m 
\end{array}\right] \in \mathbb{R}^{3m+2}
$
}
\end{equation}

\begin{equation}
\label{eq5}
SDB_i = \begin{bmatrix} 
SDB_i^1 & SDB_i^2 & \dots & SDB_i^d
\end{bmatrix} \in \mathbb{R}^d
\end{equation}

Where $i$ denotes the $i$-th observation location.

Then we construct a local relationship graph \cite{sainburg2021parametric} based on the SC by converting the differences between neighbors into weights or probabilities using Equation \eqref{eq6}.  

\begin{equation}
\label{eq6}
W_{ij} = \exp \left( -\frac{d(SC_i, SC_j) - p_i}{\sigma_i} \right)
\end{equation}

Where $d(SC_i, SC_j)$ is the distance between the $i$-th and $j$-th observation.

Apparently, a shorter distance implies a stronger spatial dependency relationship. Hence, $d(SC_i, SC_j)$ is considered as the spatial dependency between these two observations. $p_i$ is the local spatial dependency between the $i$-th observation and its adjacent neighbor. $\sigma_i$ is the local spatial dependency between the $i$-th observation and its $m$ nearest neighbors, which is regularized as $\sum_j W_{ij} = \log_2(m)$.

We then build an embedding spatial dependency basis which can preserve the structure of the local relationship graph between the nearest neighboring spatial covariates. This basis is optimized to minimize the difference between SC and SDB with respect to fuzzy set cross entropy in Equations \eqref{eq7} and \eqref{eq8}.

\begin{equation}
\label{eq7}
C(W_{ij}, \mu_{ij}) = \sum_{ij}{}\left[ W_{ij} \log \left( \frac{W_{ij}}{\mu_{ij}} \right) + (1 - W_{ij}) \log \left( \frac{1 - W_{ij}}{1 - \mu_{ij}} \right)\right]
\end{equation}

\begin{equation}
\label{eq8}
\mu_{ij} = \left[1 + a \cdot d(SDB_i, SDB_j)^{2b}\right]^{-1}
\end{equation}

The local relationship graph between extracted SDB is characterized by the weights $\mu_{ij}$. 
$d(SDB_i, SDB_j)$ denotes the distance between the $i$-th observation location and the $j$-th observation location with respect to the dataset of SDB. $a$ and $b$ are adjusted by non-linear least squares fitting.

Further implementation of UMAP, such as the fuzzy topological representation and parameter calculation, can be found in \cite{mcinnes2018umap,sainburg2021parametric}.

\subsection{Spatial dependency estimation}

Estimating a parameter field involves approximating the spatial dependency function using both the Spatial Dependency Basis (SDB) and domain knowledge, particularly in the form of automatically generated fuzzy IF-THEN rules. As illustrated in \textbf{Figure \ref{fig:network-2}}, our rule-assisted adaptive network transforms domain knowledge related to spatial dependency into a rule base.

ANFIS combines the nonlinear function approximation capabilities of neural networks with the knowledge utilization strengths of fuzzy inference systems. In this context, we adapt ANFIS to model the nonlinear spatial dependency function by generating latent knowledge from the dataset of spatial dependency bases. For example, in a 4-dimensional case, the transformed knowledge in the form of IF-THEN rules can be expressed as:

\begin{equation}
\begin{aligned}
\text{IF} \quad & SDB_{i1} \text{ is } \mathit{Rule}_{11} \text{ and } SDB_{i2} \text{ is } \mathit{Rule}_{21} \text{ and } SDB_{i3} \text{ is } \mathit{Rule}_{31} \text{ and } SDB_{i4} \text{ is } \mathit{Rule}_{41} \\
\text{THEN} \quad & \Phi_i = \overrightarrow{C_i} \cdot \overrightarrow{SDB_i}
\end{aligned}
\end{equation}

where \( \overrightarrow{SDB_i} = \left[ SDB_{i1}, SDB_{i2}, SDB_{i3}, SDB_{i4}, 1 \right] \) includes the 4-dimensional spatial dependency basis and one as a constant. \( C_i \) is the consequent parameter matrix to be discussed later, and \( \Phi_i \) is the estimated value at an unknown location. \( \mathit{Rule}_{11}, \mathit{Rule}_{21}, \mathit{Rule}_{31}, \text{and } \mathit{Rule}_{41} \) are the corresponding defined linguistic labels of each spatial dependency basis, respectively.

A complete IF-THEN rule consists of both the antecedent clause in the IF part and the consequent clause in the THEN part. Typically, linguistic labels are often defined such as “Close,” “Medium,” and “Fair.” In this study, we refer to these linguistic labels as \( \mathit{Rule}_{ij} \) for convenience. \( \mathit{Rule}_{ij} \) represents the \( j \)-th linguistic label of the \( i \)-th spatial dependency basis. Additionally, it is characterized by a bell membership function as shown in Equation \eqref{eq10}:

\begin{equation}
\label{eq10}
    \mu_{\mathit{Rule}} = \frac{1}{1 + \left[\left(\frac{s - g}{e}\right)^2 \right]^{f}}
\end{equation}

where \( s \) is the input data into the rules-set layer (shown in \textbf{Figure \ref{fig:network-2}}), \( \mathit{Rule} \) is the linguistic label, and \( e, f, g \) are premise parameters to be adjusted during the training process. The bell membership function is important as it can tolerate imprecise information by assigning a certain degree (between 0 and 1) of membership to each input \( s \) based on fuzzy logic. This quantifies the uncertainties associated with the observation data.

Within our ANFIS architecture, the rules-set layer is predefined to set the number of membership functions (or the number of linguistic labels) directly, which determines the number of the finally generated IF-THEN rules. Each node in this layer is given by Equation \eqref{eq11}:

\begin{equation}
\label{eq11}
O_\mathit{l,k}^{Rule-set} = \mu_{\mathit{Rule}_{l,k}}(x)
\end{equation}

where \( O_\mathit{l,k}^{Rule-set} \) is the \( k \)-th node or membership function of the \( l \)-th dimensional input spatial dependency basis that makes the \( SDB_l \) fuzzy between 0 and 1, where \( l = 1, 2, \dots, d \) and \( d \) is the dimensionality of the extracted spatial dependency basis. In this study, we assign two membership functions for each input, so \( k = 1, 2 \). As a result, we can write the premise parameters in \( \vec{P} = [e_{11}, f_{11}, g_{11}, e_{12}, f_{12}, g_{12}, \dots, e_{d1}, f_{d1}, g_{d1}, e_{d2}, f_{d2}, g_{d2}]_{1 \times 6d} \).

The T-Norm layer performs a multiplication operation on the signals from the previous layer. It outputs the weight \( \omega_t \) or the firing strength of a rule, according to Equation \eqref{eq12}:

\begin{equation}
\label{eq12}
    O_{\mathit{t}^{TNorm}} = \omega_t = \mu_{\mathit{Rule}_{l_1 k_1}}(s_1) \cdot \mu_{\mathit{Rule}_{l_2 k_2}}(s_2) \cdot \mu_{\mathit{Rule}_{l_3 k_3}}(s_3) \cdot \mu_{\mathit{Rule}_{l_4 k_4}}(s_4)
\end{equation}

where \( O_{\mathit{t}}^ {TNorm} \) is the \( t \)-th weight and \( t = 1, 2, 3, \dots, 2^d \), \( l_1 \neq l_2 \neq l_3 \neq l_4 \), and \( k_1, k_2, k_3, k_4 = 1, 2 \).

The normalized layer calculates the ratio of the \( t \)-th firing strength to the sum of all rules’ firing strengths using Equation \eqref{eqq13}:

\begin{equation}
\label{eqq13}
    O_{\mathit{t}}^{Normalized} =\overline{\omega_t} = \frac{\omega_t}{\sum \omega_t}, \quad t = 1, 2, 3, \dots, 2^d
\end{equation}

The consequent layer computes the result of the consequent clause of the IF-THEN rules following Equation \eqref{eqq14}. The parameters in this layer, known as consequent parameters, are adjusted during the training process using least squares estimation:

\begin{equation}
\label{eqq14}
    O_{\mathit{t}}^{Consequent} = \overline{\omega_t} f_t = \overline{\omega_t} (\overrightarrow{C} \cdot \overrightarrow{SDB}) = \overline{\omega_t} (p_1 SDB_1 + p_2 SDB_2 + \dots + p_d SDB_d + p_{d+1}) 
\end{equation}

where \( \overrightarrow{SDB} = [SDB_1, SDB_2, \dots, SDB_d, 1] \) and \( \overrightarrow{C} = [p_1, p_2, \dots, p_d, p_{d+1}] \).

The summation layer calculates the overall output by summing all the incoming signals from the previous layer, as shown in Equation \eqref{eq15}:

\begin{equation}
\label{eq15}
    O_{\mathit{summation}} = \sum O_{t}^ {Consequent} = \sum \overline{\omega_t} f_t 
\end{equation}

\subsection{Implementation of the proposed hybrid learning framework}

We outline the key implementation steps of our hybrid framework. First, we obtain the nearest neighboring spatial covariates by constructing neighboring graph for each observation point. We then use these covariates to extract the latent spatial dependency basis as the main input features for our rule-assisted adaptive networks. Next, we model and train the spatial dependency function using the rule-based ANFIS. Finally, we approximate the nonlinear spatial function, which allows us to perform spatial interpolation of a specific attribute field. Table \ref{tab:pseudocode} lists the corresponding pseudo code.

\begin{table}[ht]
\centering
\caption{Pseudo code of hybrid data-driven and rule-assisted learning procedure}
\renewcommand{\arraystretch}{1.5}
\begin{tabular}{|p{15cm}|}
\hline
\textbf{Algorithm:} Hybrid data-driven and rule-assisted learning procedure \\ \hline

\textbf{Input:} Observation record \( \mathit{Obs} = \left[\mathit{Obs}_1, \mathit{Obs}_2, \dots, \mathit{Obs}_i, \dots, \mathit{Obs}_N\right]\) \\
\textbf{Output:} Premise parameters \(\vec{P}\) and consequent parameters \(\vec{C}\) of fuzzy IF-THEN rules for establishing the approximated spatial dependency function \\ \hline

\textbf{Step 1:} Obtain nearest neighboring Spatial Covariates (\textit{SC}) for each observation point \\
\textbf{Step 1.1:} Construct neighboring graph for each observation point \\
\textbf{Function:} NeighboringGraph(\(Obs, m\)) \\
For each observation \(\mathit{Obs}_i \in Obs\): \\
\quad \# Return \(\mathit{m}\) nearest neighbors of \(\mathit{Obs}_i\) in all observations \\
\quad Neighbors $\leftarrow$ NN(\(\mathit{Obs}_i, Obs, m\)) \quad \# nearest neighbors algorithm \\
\quad NeighborsMatrix.append(Neighbors) \\
Return NeighborsMatrix \\
\textbf{Step 1.2:} Build \textit{SC} from NeighborsMatrix \\
\textbf{Step 2:} Extract \textit{SDB} from \textit{SC} using UMAP \\
\quad \# the extraction is performed by minimizing cross-entropy loss \\
\textbf{Step 3:} Model and train the spatial dependency function using rule-based ANFIS \\
\textbf{Function:} ANFIS(\(SDB, \Phi\)) \\
\textbf{Step 3.1:} Update premise parameters by the gradient descent method \\
\quad \(\vec{P} \leftarrow \vec{P} - \eta \frac{\partial(\text{error})}{\partial \vec{P}}\) \quad \# adjust premise parameters \\
\textbf{Step 3.2:} Update consequent parameters by least squares estimation \\
\quad \(\vec{C} = (SDB^T SDB)^{-1} SDB^T \Phi\) \quad \# using Kalman filtering algorithm to calculate \(\vec{C}\) \\
Return \(\vec{P}\) and \(\vec{C}\) \\ \hline
\end{tabular}
\label{tab:pseudocode}
\end{table}

\section{Applications}

To evaluate the general applicability and practical performance of the proposed framework, we apply it to two different cases involving the estimation of spatially dependent properties using interpolation of scattered observation data. The first case addresses a classic challenge in subsurface formation characterization \cite{yarus2006practical}, where the goal is to estimate the distribution of formation properties, such as rock porosity and/or permeability, based on very scattered observations, and produce a continuous property distribution map for subsurface resource exploitation. This is a particularly relevant use case as only limited observation data are available spatially in general \cite{jensen2000statistics, masoudi2024analysing}. The second case focuses on estimating the distribution of air pollutants by interpolating data collected from scattered observation stations. 

\subsection{Spatial interpolation for oil reservoir porosity mapping}

\begin{figure}[h!]
\centering
\includegraphics[width=0.9 \textwidth]{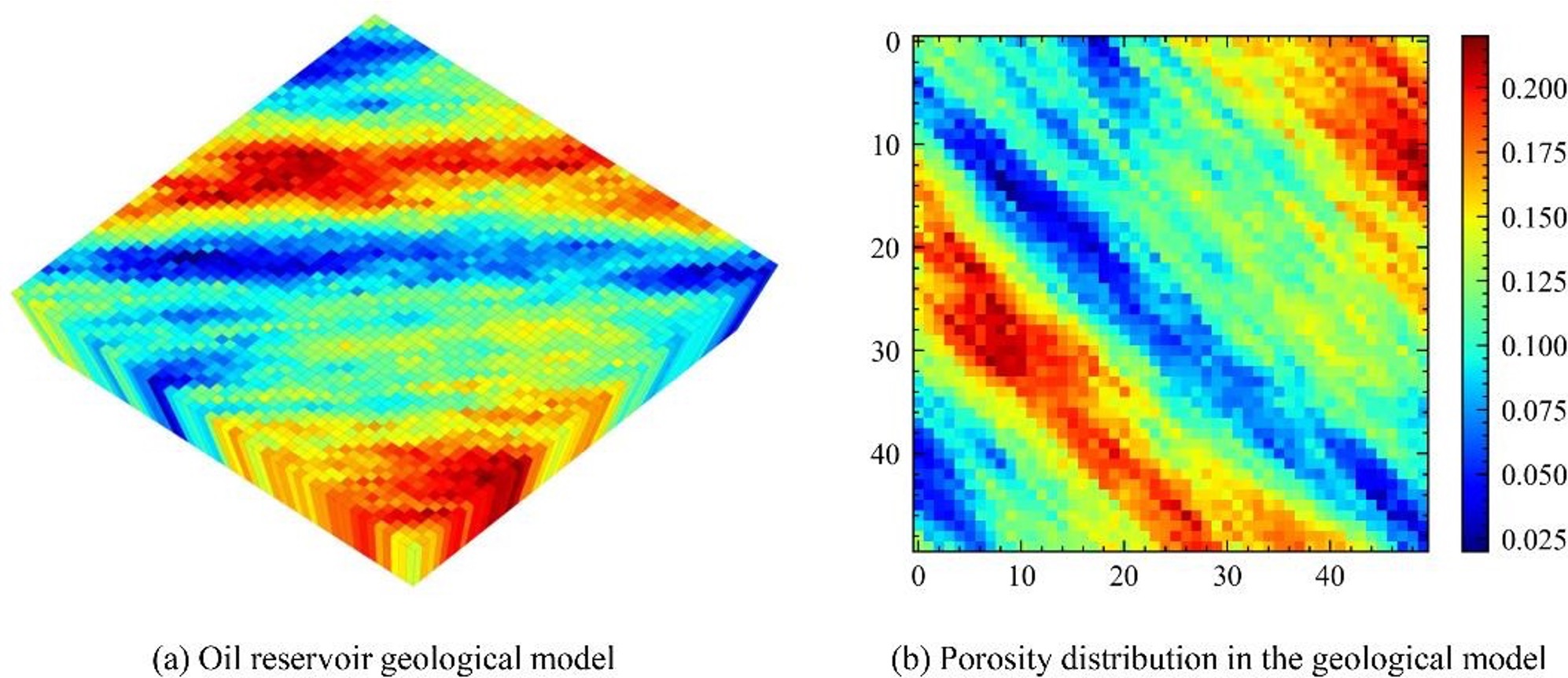}
\caption{\textbf{Oil reservoir model and its porosity distribution}}
\label{fig:resmodel}
\end{figure}

We use an oil reservoir example as a representing case. We aim to generate the formation porosity distribution using limited porosity measurements at scattered spatial observation locations. The benchmark model is provided in Figure \ref{fig:resmodel}. This model contains 50-by-50-by-1 grid blocks.    

\begin{figure}[h!]
\centering
\includegraphics[width=0.5 \textwidth]{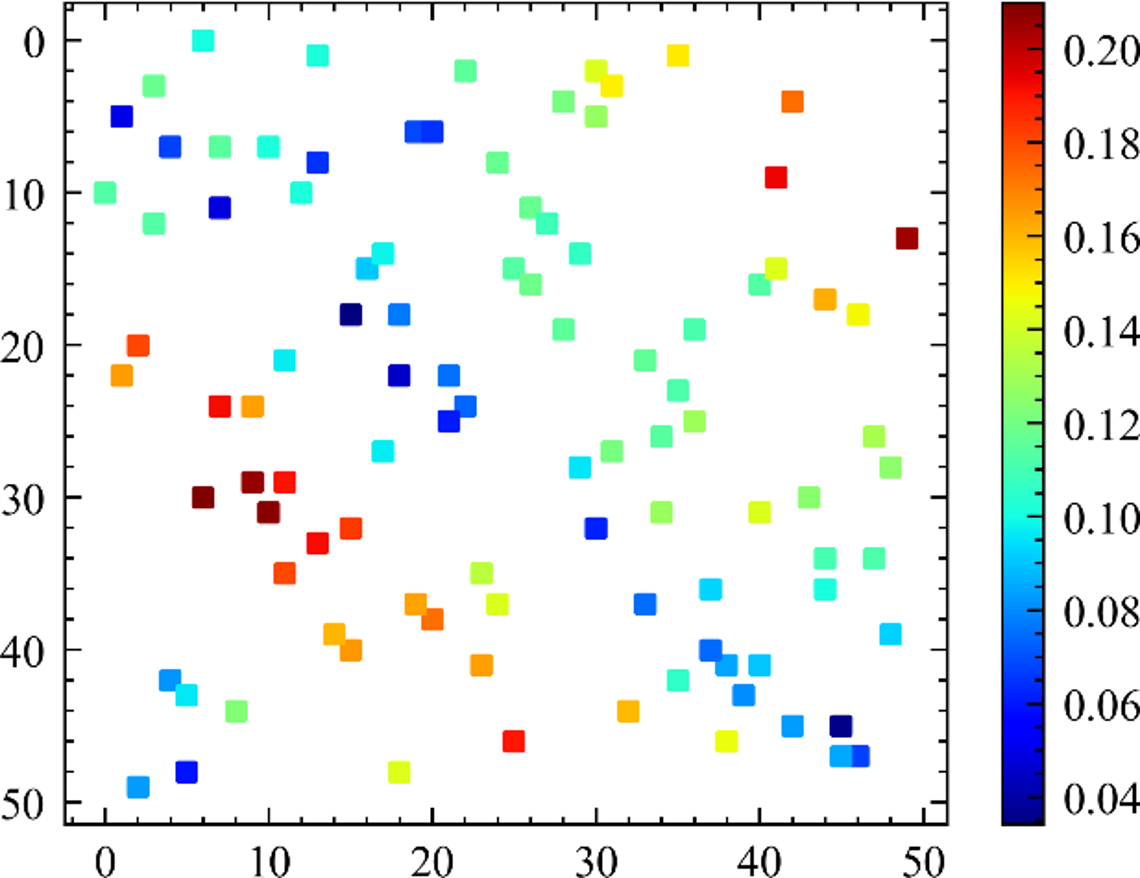}
\caption{\textbf{Scattered porosity data at randomly chosen observation locations}}
\label{fig:scatterporosity}
\end{figure}

We irregularly sample the porosity field, as shown in Figure \ref{fig:resmodel} to mimic the scattered observation locations with porosity measurement data, see Figure \ref{fig:scatterporosity}. We then randomly choose 100 observation locations as the input for the proposed hybrid learning framework. By modeling a proper spatial dependency function, we obtain the porosity estimation for unsampled locations in this model with 2500 grid blocks. 

\begin{figure}[h!]
\centering
\includegraphics[width=0.5 \textwidth]{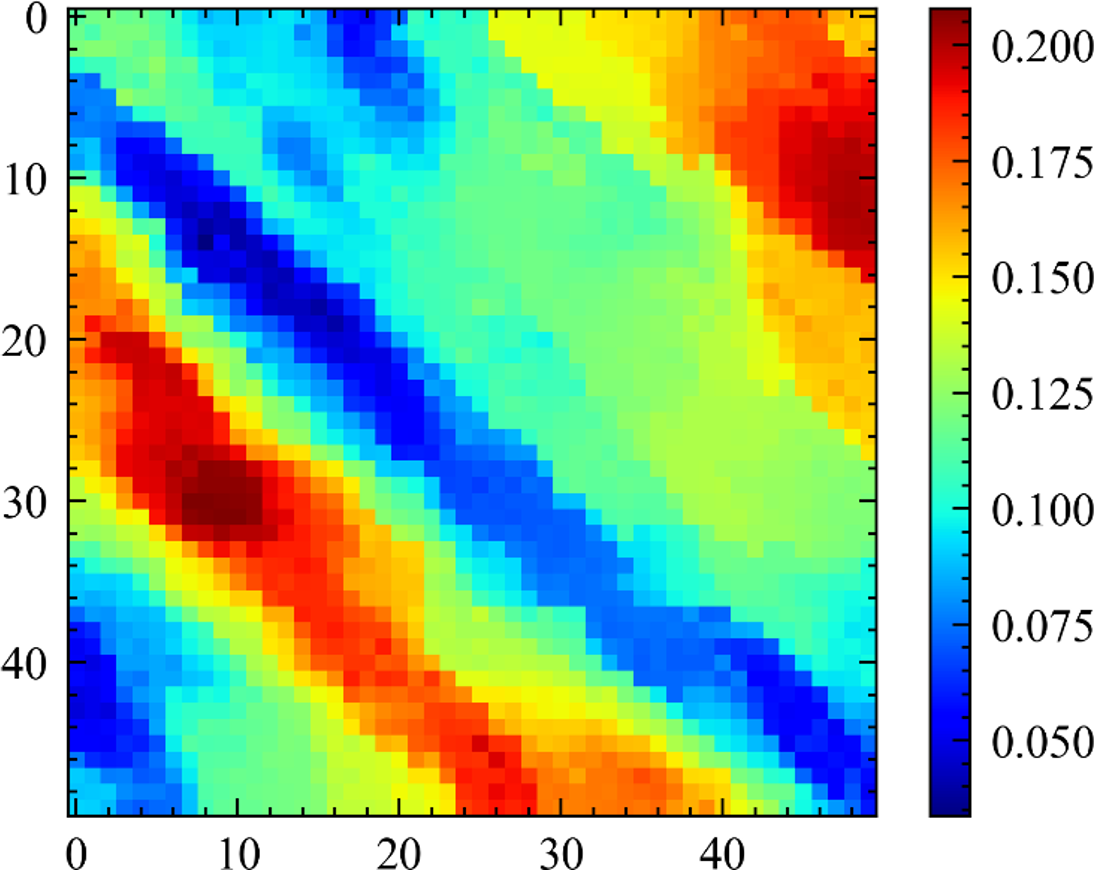}
\caption{\textbf{Porosity distribution constructed using our proposed hybrid learning framework}}
\label{fig:pormap}
\end{figure}

As shown in Figure \ref{fig:pormap}, the estimated porosity distribution map closely recaptures the original spatial pattern in the benchmark model. The high porosity channel and the low porosity channel along the lower-left diagonal are apparently reconstructed. Additionally, several dispersed regions with significant porosity variations are also qualitatively estimated, such as the high porosity region in the upper-right corner and the low porosity zone in the lower-left corner.   

\begin{figure}
    \centering
    \includegraphics[width=1\linewidth]{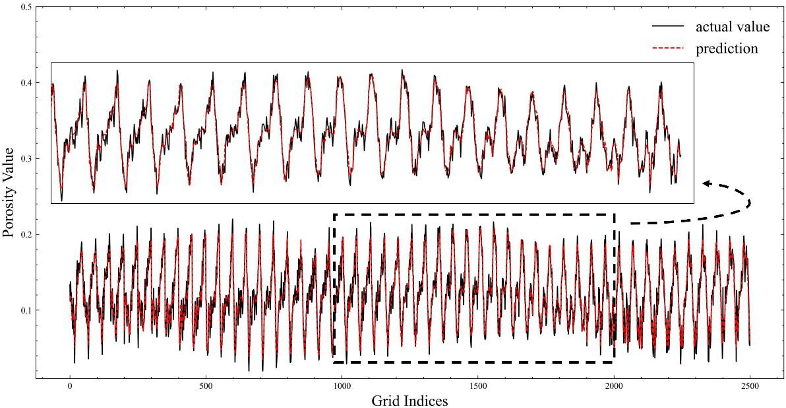}
    \caption{\textbf{comparative curve between actual value and estimated predictions for each grid}}
    \label{fig:error}
\end{figure}

We further compare the estimated porosity values with the actual benchmark values using line plots as shown in Figure \ref{fig:error}. The red dotted line represents the estimated porosity for each grid block, while the black solid line shows the actual porosity. Clearly, our estimation aligns well with the true porosity data; however, upon closer inspection of the curves, some minor discrepancies between the actual and predicted porosity values appear. To address these differences, we perform a quantitative analysis of the estimation performance using some evaluation metrics. 

The evaluation metrics include Mean Square Error (MSE), Root Mean Square Error (RMSE), Mean Absolute Error (MAE), Mean Absolute Percentage Error (MAPE), and coefficient of determination (\(R^2\)), as given in the following Equations.

\begin{align}
\label{eqq16}
\textit{MSE} &= \frac{1}{n} \sum_{i=1}^{n} (y_i - \hat{y}_i)^2 \\
\textit{RMSE} &= \sqrt{\frac{1}{n} \sum_{i=1}^{n} (y_i - \hat{y}_i)^2} \\
\textit{MAE} &= \frac{1}{n} \sum_{i=1}^{n} |y_i - \hat{y}_i| \\
\textit{MAPE} &= \frac{1}{n} \sum_{i=1}^{n} \frac{|y_i - \hat{y}_i|}{\max(y_i)} \\
R^2 &= 1 - \frac{\sum_{i=1}^{n} (y_i - \hat{y}_i)^2}{\sum_{i=1}^{n} (y_i - \bar{y})^2}
\end{align}

Where \( \hat{y}_i \) and \( y_i \) are the estimated and actual porosity of the \(i\)-th grid, respectively. \(i = 1, 2, 3, \dots, n\), and \(n\) is the total number of grid blocks in the model. The symbol \( |\cdot| \) represents the absolute value, and \( \bar{y} \) is the mean of actual porosities.

\begin{figure}
    \centering
    \includegraphics[width=0.5\linewidth]{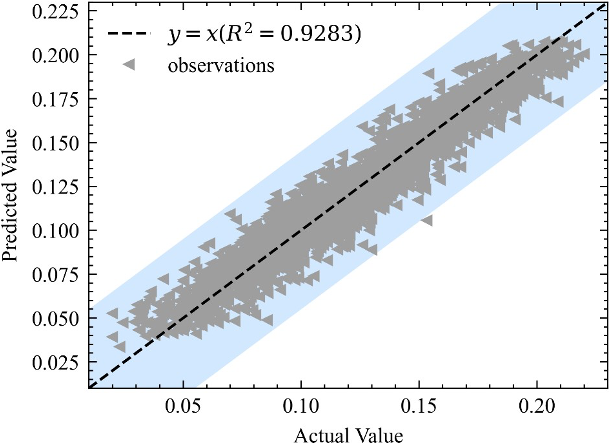}
    \caption{Interpolation performance of each grid using scatter plot along with regression straight line}
    \label{fig:r2_por}
\end{figure}

The calculated metrics are given in Table \ref{tab:evaluation_metrics1}. The errors are very low in terms of MSE, RMSE, MAE, and MAPE. The coefficient of determination (\(R^2\)) is high at 0.9283. Furthermore, we demonstrate the fitted straight line between estimated and actual porosities for a better illustration of \(R^2\) in Figure \ref{fig:r2_por}. The regression line \(y = x\) indicates that the estimation closely matches the actual porosity, underscoring the reliability of our proposed spatial interpolation method.

\begin{table}[ht]
\centering
\caption{The evaluation metrics for our proposed method}
\begin{tabular}{|c|c|c|c|c|}
\hline
\textbf{MSE} & \textbf{RMSE} & \textbf{MAE} & \textbf{MAPE} & \textbf{$R^2$} \\ \hline
0.00011236 & 0.0106 & 0.0086 & 0.0391 & 0.9283 \\ \hline
\end{tabular}
\label{tab:evaluation_metrics1}
\end{table}

\subsection{Spatial interpolation for ozone value mapping}

\begin{figure}
    \centering
    \includegraphics[width=0.6\linewidth]{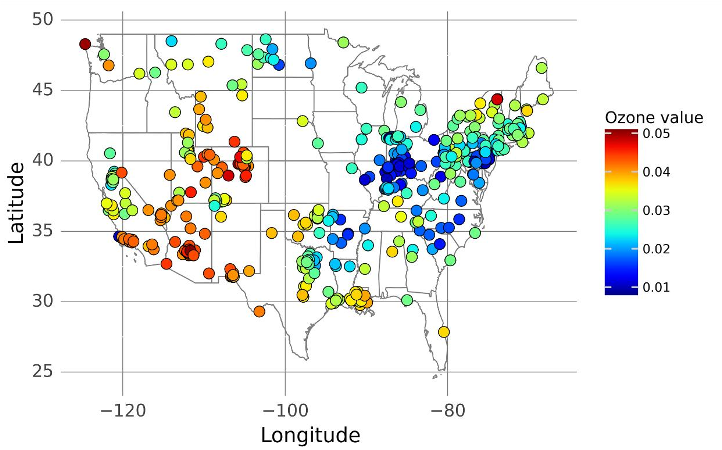}
    \caption{Ozone value at scattered monitoring station in continental United States}
    \label{fig:scatterozone}
\end{figure}

The second case is to estimate ozone value distribution based on sparse measurement data at scattered air quality monitoring stations. The ozone data used in this study is publicly available from the U.S. Environmental Protection Agency (EPA) Air Quality System (https://aqs.epa.gov/aqsweb/airdata/). Those ozone values are based on the U.S. national ambient air quality standards, specifically in the form of the “annual fourth-highest daily maximum 8-hour concentration". In our application, we select the observed ozone data recorded at the same time across various monitoring stations as our test dataset, as illustrated in Figure \ref{fig:scatterozone}. We then conduct spatial interpolation across the entire continental United States to estimate ozone levels in areas without direct observations. This allows for an assessment of air quality, even in regions with limited monitoring infrastructures. 

\begin{figure}
    \centering
    \includegraphics[width=0.6\linewidth]{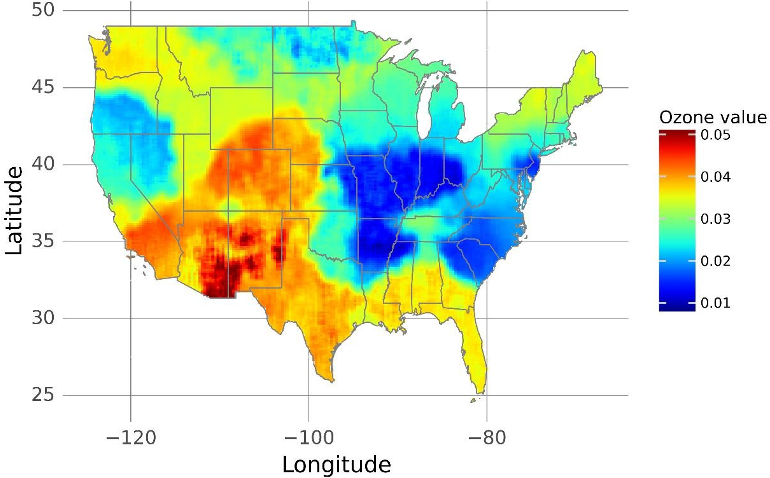}
    \caption{Ozone value distribution interpolated using our hybrid framework}
    \label{fig:ozonemap}
\end{figure}

The map generated using our proposed method captures the higher ozone areas in the west and lower areas in the east, as clearly and intuitively visualized in Figure \ref{fig:ozonemap}. The reconstructed ozone distribution demonstrates the capabilities in capturing spatial variations, effectively distinguishing between highly contaminated regions and less affected areas. This is of practical use for representing the whole air quality situation in sparsely monitored regions, where fewer monitoring stations are available. The ability to reflect these variations in regions with limited data highlights the robustness and practical utility of our spatial interpolation method for air quality monitoring and assessment.  

In addition to visualizing the spatial distribution patterns of ozone levels, we also provide quantitative results using the evaluation metrics mentioned earlier. Unlike the first case, where actual values at unknown locations are available for comparison, the actual ozone values at non-monitoring stations in this scenario are not accessible. Therefore, we cannot directly compare the estimated values with actual measurements at those locations. We employ the 10-fold cross-validation to conduct the comparison. We divide the observed data into a training dataset and a validation dataset to assess the model's performance. We then calculate the average values for MSE, RMSE, MAE, MAPE, and \(R^2\) across the 10 validation datasets. The results of these evaluations are listed in Table \ref{tab:evaluation_metrics2}. 

\begin{table}[ht]
\centering
\caption{The evaluation metrics of 10-fold validation \\ datasets using our proposed method}
\begin{tabular}{|c|c|c|c|c|}
\hline
\textbf{MSE} & \textbf{RMSE} & \textbf{MAE} & \textbf{MAPE} & \textbf{$R^2$} \\ \hline
0.00002601 & 0.0051 & 0.0037 & 0.0792 & 0.7626 \\ \hline
\end{tabular}
\label{tab:evaluation_metrics2}
\end{table}

\section{Comparison and Discussion}

We compare our hybrid framework with several established interpolation methods, including Inverse Distance Weight (IDW) \cite{shepard1968two}, ordinary kriging \cite{matheron1963principles}, and gaussian process \cite{quinonero2005unifying}. These three comparative methods are widely used for spatial interpolation. 

\subsection{Comparison in terms of evaluation metrics}

Table \ref{tab:por_comparison} and Table \ref{tab:ozone_comparison} provide the respective comparative results. Our proposed method outperforms in terms of MSE, RMSE, MAE, MAPE, and \(R^2\) for both cases.    

\begin{table}[ht]
\centering
\caption{Comparative evaluation metrics of different interpolation methods for reservoir porosity mapping}
\label{tab:por_comparison}
\renewcommand{\arraystretch}{1.3}  % Adjust row height
\begin{tabular}{|c|c|c|c|c|}
\hline
\textbf{Metrics} & \textbf{Ordinary Kriging} & \textbf{IDW} & \textbf{Gaussian Process} & \textbf{Our Method} \\ \hline
\textbf{MSE}   & 0.00046976 & 0.00113792 & 0.00030674 & 0.00011236 \\ \hline
\textbf{RMSE}  & 0.02167388 & 0.03373304 & 0.03373304 & 0.0106 \\ \hline
\textbf{MAE}   & 0.01617806 & 0.02701021 & 0.01324166 & 0.0086 \\ \hline
\textbf{MAPE}  & 0.15395796 & 0.2807828  & 0.13766530 & 0.0391 \\ \hline
\textbf{$R^2$} & 0.71519912 & 0.31011110 & 0.81403229 & 0.9283 \\ \hline
\end{tabular}
\end{table}

\begin{table}[ht]
\centering
\caption{Comparative evaluation metrics of different interpolation methods for ozone value mapping}
\label{tab:ozone_comparison}
\renewcommand{\arraystretch}{1.3}  % Adjust row height
\begin{tabular}{|c|c|c|c|c|}
\hline
\textbf{Metrics} & \textbf{Ordinary Kriging} & \textbf{IDW} & \textbf{Gaussian Process} & \textbf{Our Method} \\ \hline
\textbf{MSE}   & 0.00003324 & 0.00002602 & 0.00002665 & 0.00002601 \\ \hline
\textbf{RMSE}  & 0.00576572 & 0.00510121 & 0.00516196 & 0.0051 \\ \hline
\textbf{MAE}   & 0.00447331 & 0.00411962 & 0.00411051 & 0.0037 \\ \hline
\textbf{MAPE}  & 0.13360340 & 0.12824334 & 0.12676602 & 0.0792 \\ \hline
\textbf{$R^2$} & 0.29874051 & 0.45106761 & 0.43791603 & 0.7626 \\ \hline
\end{tabular}
\end{table}

\subsection{Comparison in terms of spatial feature reconstruction}

\begin{figure}
    \centering
    \includegraphics[width=0.8\linewidth]{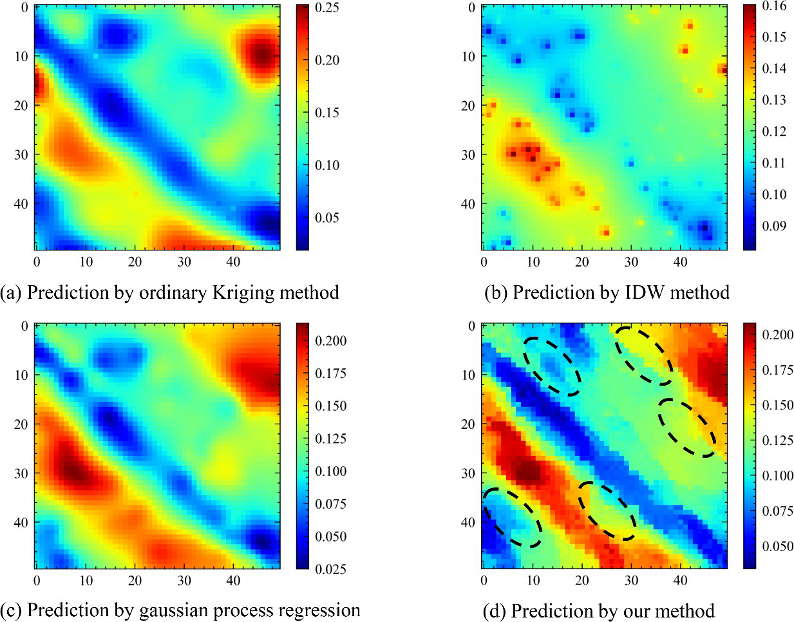}
    \caption{Comparative results of porosity interpolation generated from different methods}
    \label{fig:por_compare}
\end{figure}

In addition to quantitative analysis in terms of error and accuracy, we also compare the spatial distribution patterns from both global and local perspectives. The comparative spatial distribution maps of porosity are shown in Figure \ref{fig:por_compare}. It is evident that all three traditional methods can reconstruct the global trends and features. Global features here mean the prominent spatial distribution patterns that are easily recognizable across the entire map. These global patterns include the high porosity channel or region in the lower-left diagonal and upper-right direction, as well as the low porosity channel along the diagonal. 

However, when it comes to capturing spatially local features, all three traditional methods fall short. Local features refer to the finer details and variations within smaller regions that contribute to the overall spatial heterogeneity. The inability of these methods to accurately reconstruct these local features highlights a limitation in their capacity to fully represent complex spatial variability. This comparison underscores the importance of methods that can address both global and local spatial patterns to provide a more comprehensive and detailed understanding of the spatial distribution. As demonstrated in Figure \ref{fig:por_compare}(d), the circled regions highlight the local features that reflect variations within the porosity field. Compared to other commonly used techniques, our proposed method captures and represents these local features more effectively, providing a more detailed and nuanced depiction of the spatial variability. This enhanced ability to preserve local variations underscores the strength of our approach in handling complex spatial patterns.

\begin{figure}
    \centering
    \includegraphics[width=0.8\linewidth]{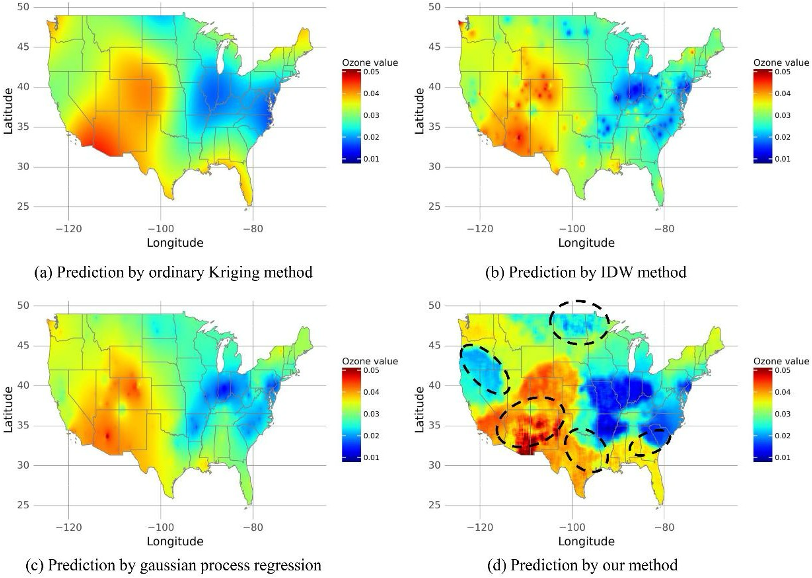}
    \caption{Comparative results of ozone interpolation generated from different methods}
    \label{fig:ozone_compare}
\end{figure}

Similarly, the comparative spatial distribution maps of ozone values are shown in Figure \ref{fig:ozone_compare}. All methods can predict global trends to some extent. For instance, the highly polluted region in the southwest and the less polluted region in the east are consistently identified across all methods. However, when it comes to reconstructing local features, our proposed method outperforms the other three interpolation techniques, as indicated by the dashed circles. 

It should be noted that the three commonly used interpolation techniques tend to produce overly smooth interpolations. While they may capture general trends, it often comes at the cost of losing important local variations within the spatial field. Excessively smooth interpolation can negatively impact the accuracy of spatial predictions by overlooking these critical local features or variations in the spatial dependency field. Our proposed method, by contrast, strikes a better balance between capturing global trends and preserving local details, leading to more accurate and nuanced spatial interpolations.

\subsection{Sensitive analysis with respect to the number of nearest neighbors}

Although our proposed method offers advantages over traditional techniques in terms of accuracy and the preservation of local features in both cases, its performance in interpolating ozone is slightly less effective compared to its success in reconstructing the porosity field. The primary difference between these two cases is the density of observed data within the interpolation space. The denser data in the porosity field allows for more precise predictions, whereas the sparser data in the ozone case makes it more challenging to capture local variations accurately. 

Compared to the first porosity field case, the observation density in the ozone case is less uniform than in our randomly selected porosity data. This nonuniform observation density tends to be more sensitive to the number of nearest neighbors \textit{m} when constructing nearest neighboring spatial covariates in our proposed method. In other words, the choice of the number of nearest neighbors has certain impact on the spatial distribution patterns and the estimation accuracy of the reconstructed attribute fields. This sensitivity underscores the importance of carefully selecting the number of nearest neighbors to ensure high quality spatial interpolation, particularly in cases with uneven observation densities. 

\begin{figure}
    \centering
    \includegraphics[width=0.8\linewidth]{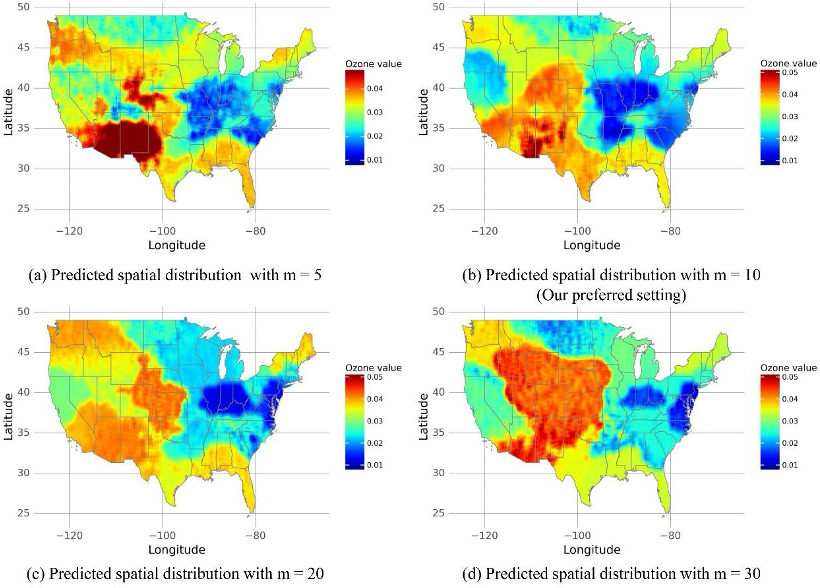}
    \caption{Reconstructed spatial distribution patterns with different number of nearest neighbors}
    \label{fig:ozone_sens}
\end{figure}

To further illustrate this point, we focus on the ozone case to discuss the sensitivity of the number of nearest neighbors \textit{m}. We compare and analyze different hyperparameter settings, as demonstrated in Figure \ref{fig:ozone_sens}. This comparison highlights how varying the number of nearest neighbors affects the spatial interpolation results, allowing us to better understand the impact of this parameter on the accuracy and reliability of our estimated predictions.

Clearly, different values of \textit{m} have significant impact on the interpolated spatial distribution and its corresponding accuracy. We choose to use 10 nearest neighbors to achieve the desired accuracy. However, it is important to note that the optimal value of \textit{m} is highly dependent on the specific observed sampling data. Considerable time and effort are required to carefully tune the hyperparameter \textit{m} to achieve higher interpolation quality. This tuning process is crucial for adapting the model to different datasets and ensuring reliable interpolation results.

Additionally, a larger value of \textit{m} tends to preserve global spatial patterns, while a smaller value of \textit{m} tends to emphasize local features. For instance, when \textit{m} is set to 30, the estimated high-ozone value region in the middle appears more expansive. Conversely, when \textit{m} is reduced to 5, the results reveal some sparse low-ozone value areas within the high-ozone value region. This occurs because a larger \textit{m} means that a broader neighborhood is considered when estimating unknown locations, leading to results that are more influenced by higher attribute values within that neighborhood. On the other hand, a smaller \textit{m} often overlooks the general trend due to its limited neighborhood size, resulting in a focus on local variations. This, however, makes the interpolations more susceptible to outliers or extreme values. 

Therefore, neither a large nor a small value of \textit{m} is ideal for constructing nearest neighboring spatial covariates. Careful tuning of the hyperparameter \textit{m} is essential when implementing our hybrid data-driven and rule-assisted learning framework for spatial interpolation. Finding the right balance in \textit{m} allows for a better representation of both global and local patterns, ensuring reliable and robust estimations.

\section{Conclusions}

In this study, we develop a hybrid data-driven and rule-assisted learning framework that combines spatial feature extraction with domain knowledge integration via fuzzy rule sets to enhance the interpolation of spatially dependent properties. By decomposing nearest neighbor spatial covariates into multiple spatial dependency bases and utilizing fuzzy IF-THEN rules within an adaptive network framework, our method accommodates imprecise information, considers data uncertainties, and improves spatial interpolation accuracy. Validated through applications in subsurface formation characterization and air quality assessment, our approach surpasses traditional techniques like ordinary Kriging, inverse distance weighting, and Gaussian processes by achieving lower error metrics and better capturing local spatial features. However, the method’s performance is sensitive to the number of nearest neighbors used, which influences the balance between global trends and local variations. While our approach is parametric and requires careful tuning, future research could explore non-parametric methods that dynamically adjust the number of nearest neighbors, offering a promising avenue for further improving spatial interpolation techniques.

\bibliographystyle{unsrt}
\bibliography{interpolation}

\end{document}